\documentclass{article}
\usepackage{ijcai25}

\usepackage{times}
\usepackage{soul}
\usepackage{url}
\usepackage[hidelinks]{hyperref}
\usepackage[utf8]{inputenc}
\usepackage[small]{caption}
\usepackage{graphicx}
\usepackage{amsmath}
\usepackage{amsthm}
\usepackage{booktabs}
\usepackage{algorithm}
\usepackage{algorithmic}
\usepackage[switch]{lineno}
\usepackage{xcolor}
\usepackage{amssymb}
\usepackage{multirow}

\title{Asymmetric Co-Training for Source-Free Few-Shot Domain Adaptation}

\author{
Gengxu Li$^1$
\and
Yuan Wu$^1$\footnote{Corresponding author}\\
\affiliations
$^1$School of Artificial intelligence, Jilin University\\
\emails
yuanwu@jlu.edu.cn
}

\begin{document}

\maketitle

\begin{abstract}
    Source-free unsupervised domain adaptation (SFUDA) has gained significant attention as an alternative to traditional unsupervised domain adaptation (UDA), which relies on the constant availability of labeled source data. However, SFUDA approaches come with inherent limitations that are frequently overlooked. These challenges include performance degradation when the unlabeled target data fails to meet critical assumptions, such as having a closed-set label distribution identical to that of the source domain, or when sufficient unlabeled target data is unavailable—a common situation in real-world applications. To address these issues, we propose an asymmetric co-training (ACT) method specifically designed for the SFFSDA scenario. SFFSDA presents a more practical alternative to SFUDA, as gathering a few labeled target instances is more feasible than acquiring large volumes of unlabeled target data in many real-world contexts. Our ACT method begins by employing a weak-strong augmentation to enhance data diversity. Then we use a two-step optimization process to train the target model. In the first step, we optimize the label smoothing cross-entropy loss, the entropy of the class-conditional distribution, and the reverse-entropy loss to bolster the model's discriminative ability while mitigating overfitting. The second step focuses on reducing redundancy in the output space by minimizing classifier determinacy disparity. Extensive experiments across four benchmarks demonstrate the superiority of our ACT approach, which outperforms state-of-the-art SFUDA methods and transfer learning techniques. Our findings suggest that adapting a source pre-trained model using only a small amount of labeled target data offers a practical and dependable solution.
    The code is available at \url{https://github.com/gengxuli/ACT}.
\end{abstract}

\section{Introduction}

In recent years, deep neural networks (DNNs) have achieved remarkable success across a wide array of applications. However, these achievements often rest on a key assumption: the training data (source domain) and test data (target domain) are drawn from the same distribution. In real-world scenarios, this assumption frequently does not hold~\cite{goodfellow2016deep}. When the test data distribution diverges from that of the training data, DNN models suffer a substantial performance decline~\cite{buda2018systematic}. This phenomenon, known as domain shift~\cite{ben2010theory}, limits the generalizability of DNNs. Unsupervised Domain Adaptation (UDA)~\cite{sun2016return,long2018conditional,wu2020dual,du2021cross} has emerged as a promising solution to address domain shift by aligning learned representations between source and target domains, thereby improving model performance. Nevertheless, UDA's fundamental assumption—that both source and target data are simultaneously available during adaptation—is often violated due to constraints like data privacy, storage costs, transmission expenses, or the computational load of training on large-scale source data~\cite{fang2022source}.

\begin{figure}
    \centering
    \includegraphics[width=0.9\columnwidth]{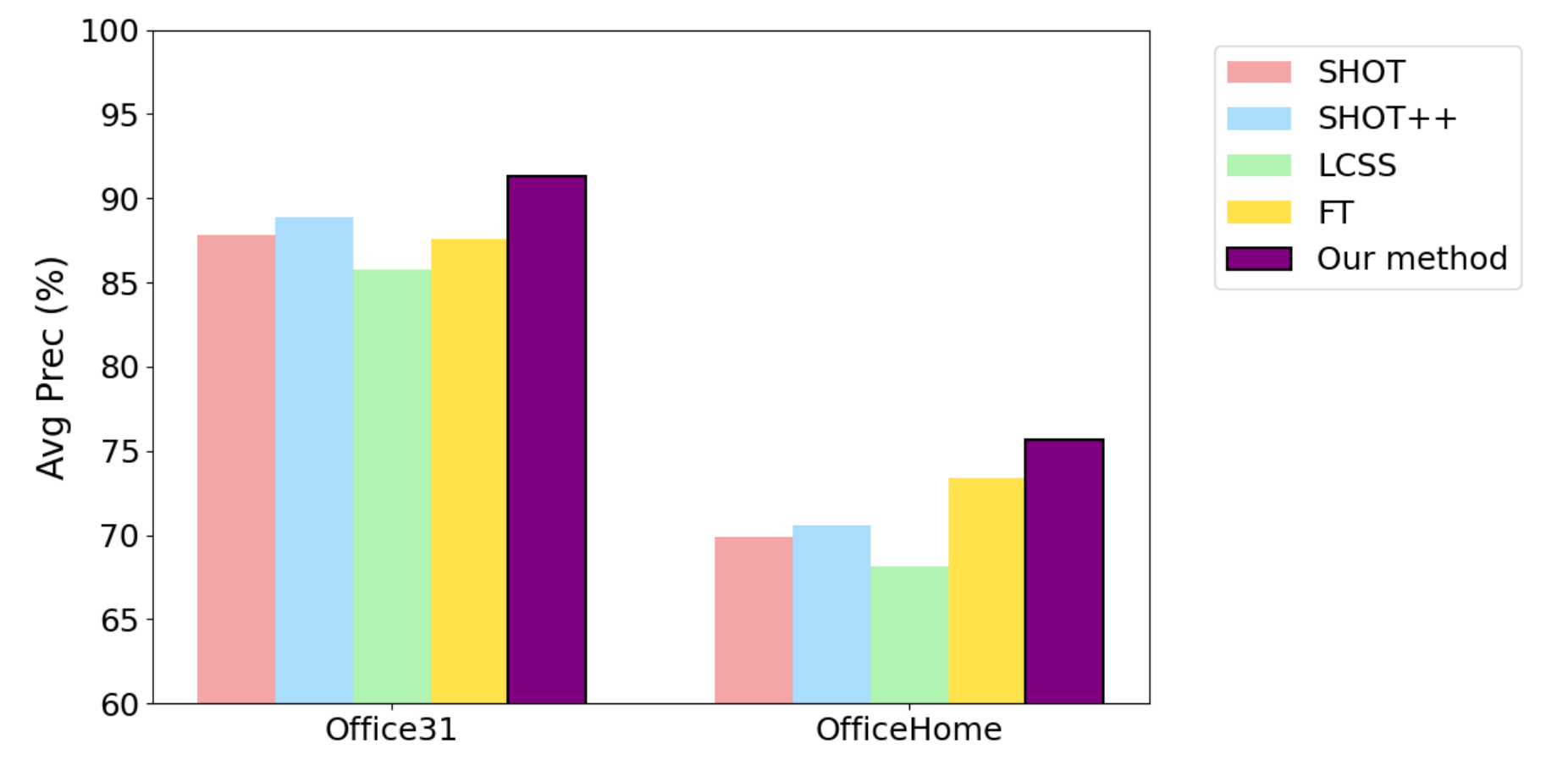}
    \caption{Comparison with SFUDA and transfer learning methods on OfficeHome dataset under 3-shot setting.}
    \label{Fig1}
\end{figure}

To address these limitations, source-free unsupervised domain adaptation (SFUDA)~\cite{liang2021source,qiu2021source,tian2021vdm,liang2020really} has been introduced. The primary distinction between UDA and SFUDA lies in the availability of source data. In SFUDA, source domain data is inaccessible during the adaptation process, and instead, the adaptation is based on a source pre-trained model, aligned with large-scale unlabeled target data~\cite{liang2020really}. Due to growing concerns about data privacy, intellectual property, and the shift towards large-scale pre-training, the source-free paradigm has garnered considerable attention from the machine learning community~\cite{li2024comprehensive}. However, SFUDA models, which depend entirely on the large-scale pre-provided unlabeled target data, face several challenges in real-world applications. First, sufficient unlabeled target data is not always available, complicating model adaptation with limited data. Second, the closed-set assumption, which assumes all target data falls within the source domain's classes, does not always hold. In practice, unlabeled target data may contain out-of-distribution (OOD) classes, resulting in adaptation failure. Finally, a shift in label distribution between source and target domains can further undermine adaptation. For example, a major class in the source domain might become a minor class in the target domain, disrupting the transfer of knowledge from the pre-trained model.

In this work, We focus on source-free few-shot domain adaptation (SFFSDA), a novel setting where a small labeled support set ($N \times K$ instances) from the target domain is used to adapt pre-trained source models. Inspired by the $N$-way $K$-shot paradigm, SFFSDA is cost-effective and bypasses the need for large amounts of unlabeled target data, offering a practical alternative to SFUDA. Previous transfer learning works, like LCCS~\cite{zhang2022few}, also try to address SFFSDA problems by using test-time adaptation. Unlike test-time adaptation methods relying on streaming data and affected by mini-batch size or class distribution, SFFSDA adapts using a small, labeled support set. LCCS reduces adaptation parameters via a low-dimensional approximation of batch normalization statistics, enabling adaptation with just one sample per class and achieving strong performance across benchmarks. However, these methods still encounter some issues, such as requiring large amounts of training data and high-quality few-shot samples. To tackle these issues, we propose an asymmetric co-training (ACT) method, which uses strong-weak augmentation to diversify the support set and a two-step optimization process to adapt pre-trained source models. In the optimization process, we first optimize label smoothing cross-entropy loss, class-conditional entropies, and reverse cross-entropy loss to enhance model's discriminability. Second, we minimize classifier determinacy disparity to refine predictions. Additionally, Sharpness-Aware Minimization (SAM) is incorporated to improve the system performance. Experiments on multiple benchmarks demonstrate ACT’s superiority over state-of-the-art SFUDA methods, including SHOT~\cite{liang2020really}, SHOT++\cite{liang2021source}, LCCS\cite{zhang2022few}, FT, and LP-FT~\cite{lee2023fewshot}. While SFFSDA operates without access to source data and relies on minimal labeled target examples, making it highly practical for real-world applications, it fundamentally differs from SFUDA, which requires large amounts of unlabeled target data. This distinction highlights the practical advantages of SFFSDA in scenarios where a limited amount of labeled target data can be obtained, yet it also necessitates reliable comparisons with unsupervised methods.


\section{Related Work}

\subsection{Unsupervised Domain Adaptation (UDA)} 

Unsupervised Domain Adaptation (UDA) methods address domain shift by jointly training a network on labeled source data and unlabeled target data, thereby aligning their feature distributions. Conventional UDA approaches tackle domain shift through two main strategies. Some methods~\cite{long2015learning,sun2016return,tzeng2014deep} employ moment matching to align feature distributions, with~\cite{tzeng2014deep} being the first to introduce Maximum Mean Discrepancy (MMD) in domain adaptation. Others~\cite{ganin2015unsupervised,long2018conditional,wu2020dual} utilize adversarial training to learn domain-invariant features. Additionally, certain methods~\cite{saito2018maximum,li2021bi,du2021cross} apply adversarial training between the feature extractor and the classifier. These techniques decouple the source and target domains during training, allowing the model to estimate the domain difference without direct access to the source data. However, under the UDA setting, the target labels remain inaccessible during training. As a result, UDA methods often rely heavily on the accuracy of prototype estimation and pseudo-label annotation for effective adaptation.

\subsection{Source-Free Unsupervised Domain Adaptation (SFUDA)}

Conventional UDA methods rely on supervision from the source domain during the adaptation process, which can be challenging to achieve in practical settings. As a result, source-free unsupervised domain adaptation (SFUDA) methods have gained significant attention, as they aim to adapt models to the target domain without accessing source domain data. From the DA perspective, works such as~\cite{kundu2020universal,kundu2020towards} have explored source-free universal DA~\cite{you2019universal} and open-set DA~\cite{saito2018open}. SHOT~\cite{liang2020really}, the pioneering SFUDA method, primarily addresses closed-set DA and partial-set DA~\cite{li2020deep}. To achieve SFUDA, certain methods~\cite{qiu2021source,tian2021vdm} focus on reconstructing a pseudo-source distribution in feature space based on the source hypothesis, enhancing generalization by aligning target domain samples with these pseudo-source samples. Another prominent stream of SFUDA methods~\cite{liang2020really,chen2022self} concentrates on pseudo-label prediction from the source model or prototypes, adapting the model to the target domain and ensuring a better fit to the target domain distribution.

\subsection{Few-Shot Transfer Learning}

A wide array of studies focus on learning a metric space specifically for $k$-shot tasks, leveraging meta-learning to develop adaptation strategies~\cite{pan2009survey}. These approaches, however, often require specialized network architectures, custom loss functions, and training strategies that depend on multiple source domains or joint training with both source and support data—conditions that are challenging to meet in real-world environments. Another commonly adopted strategy is model fine-tuning or weight transfer from a pre-trained source model. However, fine-tuning all source model weights on a limited support set with a small value of $k$ typically leads to severe overfitting, necessitating a support set of at least $k = 100$ to achieve stability~\cite{yosinski2014transferable}. Recent advancements have been made in the SFFSDA setting. For example, LCCS~\cite{zhang2022few} utilizes a low-dimensional approximation of batch normalization (BN) statistics to drastically reduce the number of parameters required for adaptation, thereby improving the robustness of adaptation with limited support data. Additionally, ~\cite{lee2023fewshot} proposes that fine-tuning a source pre-trained model using a few labeled target instances presents a more feasible solution to overcome the limitations inherent in SFUDA, and they successfully demonstrate the effectiveness of this approach. In contrary to prior works, our proposed ACT method can not only effectively tackle closed-set SFFSDA problem, but also tackle open-set and partial-set domain adaptation scenarios.  

\section{Method} \label{sec:three}

\subsection{Overall Idea} \label{sec:three.one}

First, let us describe our experimental setup. We denote the source model, pre-trained on the labeled source domain $\{X_S,Y_S\}$, as $\Phi_S$, where $X_S$ represents the source data and $Y_S$ denotes the corresponding label information. Similarly, $\{X_T,Y_T\}$ represents the labeled target domain, consisting of target data and their associated labels. The objective of SFFSDA is to sample a support set $\{X_{SUP},Y_{SUP}\}$ using the few-shot approach, specifically the $N$-way-$K$-shot setting, where $N$ is the number of target domain classes and $K$ is the number of samples drawn from each class. Using the labeled data $\{X_{SUP},Y_{SUP}\}$ from the support set, we fine-tune the pre-trained source model $\Phi_S$ to learn a target model $\Phi_T$. In the SFFSDA scenario, we cannot access the original source data (i.e., $X_S$ and $Y_S$), but we do have access to the parameters of the source model $\Phi_S$. During domain adaptation, only the support set data $\{X_{SUP},Y_{SUP}\}$ is available, and this data is used to fine-tune the pre-trained source model $\Phi_S$.

In contrast to existing SFFSDA method~\cite{lee2023fewshot}, which adapts source pre-trained models via fine-tuning the target support data, we propose a novel approach. We train a feature extractor network $F$, which is initially trained on the source inputs $X_S$ and later fine-tuned with the support inputs $X_{SUP}$. Additionally, we employ two classifier networks $C_1$ and $C_2$, which use the features generated by $F$. Both classifiers map the features into $K$ classes, outputting a $K$-dimensional vector. A softmax function is applied to these vectors to obtain class probabilities, denoted as $p_1(y|x)$ and $p_2(y|x)$, representing the probability outputs from classifiers  $C_1$ and $C_2$, respectively.

Unlike traditional SFUDA methods, SFFSDA allows access to target domain labels. Therefore, methods such as pseudo-labeling and self-training, commonly used in SFUDA, are not required. However, since the labeled target data in the support set is typically very limited due to the few-shot setting, the lack of data diversity can lead to overfitting. Addressing this issue is crucial to enhancing the effectiveness of SFFSDA methods. Given the source-free setting, where access to the source data is restricted, it is essential to establish a strong relationship between the pre-trained source model and the target model. Since our approach incorporates two classifiers, minimizing the discrepancy between the outputs of both classifiers is vital for the model's performance. Based on these principles, we have designed our method.

\subsection{Data Diversity} \label{sec:three.two}

Due to the few-shot setting, the support set used to fine-tune the pre-trained model typically contains limited data, leading to potential issues like overfitting and lack of diversity. To address these challenges and enhance data variety, we use a strong-weak data augmentation strategy inspired by contrastive learning. Specifically, we apply two augmentation methods, ${A_S,A_W}$: the weak augmentation $A_W$, which involves basic transformations like random cropping and flipping, and the strong augmentation $A_S$, which uses more complex techniques through AutoAugment~\cite{cubuk2019autoaugment}, an automated method that optimizes augmentation strategies using search algorithms. AutoAugment combines 24 sub-strategies to identify the most effective augmentations for the dataset. These strategies generate synthetic data from ground-truth targets, which is particularly useful in few-shot settings. In this study, augmentations serve to enhance data diversity, thereby mitigating the risk of overfitting.


For our experiments, we construct the support set adhering to the few-shot learning framework, specifically employing an $N$-way-$K$-shot setting. Unlike previous SFFSDA work~\cite{lee2023fewshot}, which relied on less rigorous data-splitting strategies, we have refined this process for greater scientific integrity. Rather than arbitrarily partitioning the target data, we systematically select samples from the entire target dataset to form the support set in accordance with few-shot principles. The remainder of the target data, not included in the support set, is designated as the test set, thereby eliminating the need for a separate validation set. This approach ensures that our experimental setup is both methodologically sound and aligned with the few-shot learning paradigm.

By employing these augmentation strategies, we inject controlled randomness into the support set, effectively increasing the diversity and effective volume of the available data. Our refined data-splitting method ensures that the test set remains representative and diverse, while the support set benefits from a well-maintained level of randomness. This approach not only enhances the testing accuracy but also leads to improved model performance, ensuring robust evaluation and reliable outcomes.


\subsection{Model Structure} \label{sec:three.three}

\begin{figure}[!h]
    \centering
    \includegraphics[width=1.0\columnwidth]{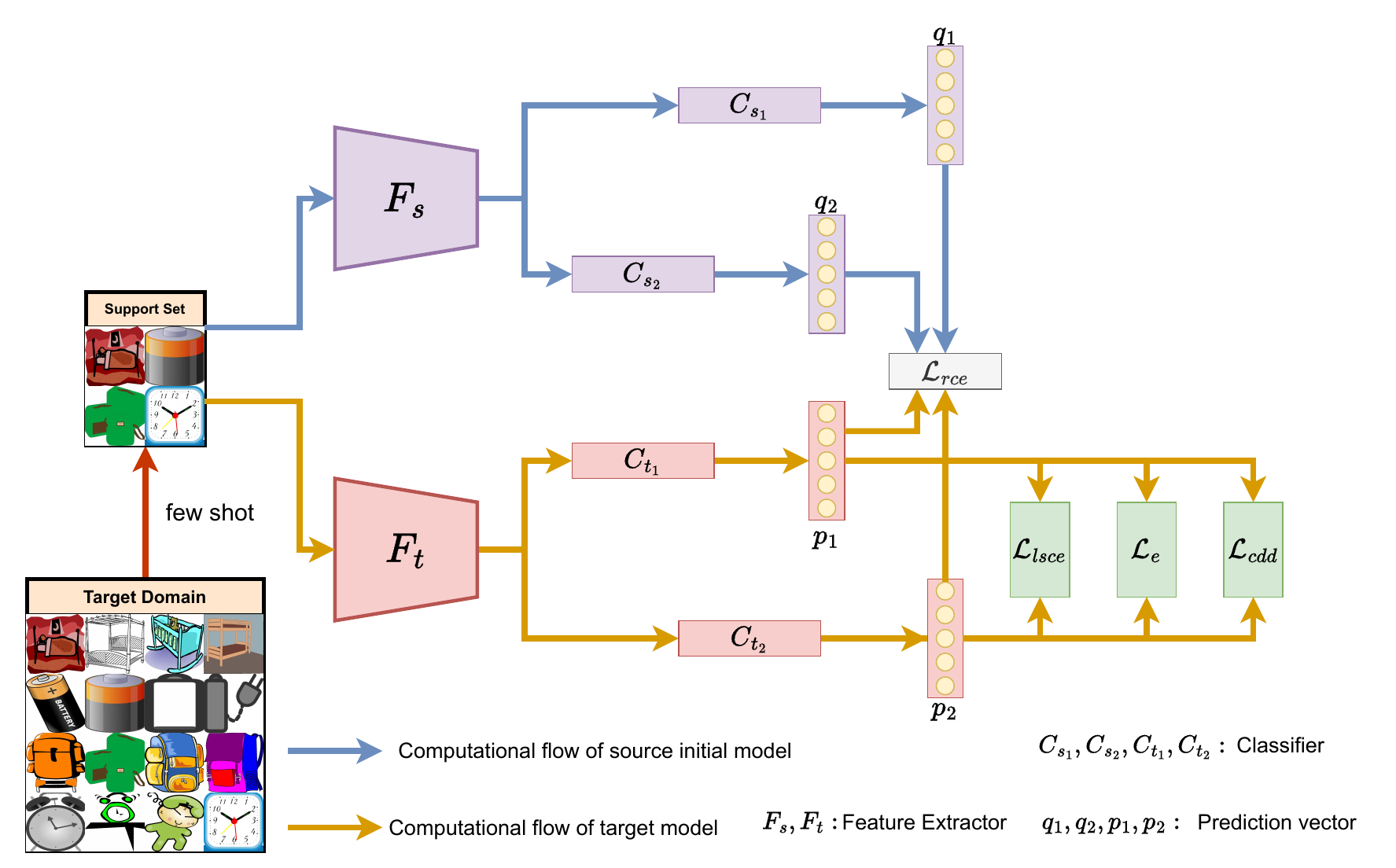}
    \caption{Overview of the ACT model structure and objective functions.}
    \label{Fig2}
\end{figure}

Our proposed ACT model contains two modules, each comprising a feature extractor and two classifiers. In the source-free setting, the absence of labeled source domain samples during training requires an alternative approach to obtain pre-trained source features. To achieve this, we first train a feature extractor $F_{src}$ and two classifiers $C_{{src}_1}$ and $C_{{src}_2}$ on the source domain. The modules $F_s$, $C_{s_1}$, and $C_{s_2}$ are then initialized with the parameters of $F_{src}$, $C_{{src}_1}$, and $C_{{src}_2}$, respectively, serving as a substitute for the unavailable labeled source data. These modules remain frozen during target domain fine-tuning to retain the source pre-trained features. Our method aims to train the target model $F_t$, $C_{t_1}$, and $C_{t_2}$, they are also initialized by using the parameters of $F_{src}$, $C_{{src}_1}$, and $C_{{src}_2}$, by using a support set sampled from the target domain. Figure~\ref{Fig2} illustrates the model architecture.

\subsection{Training Steps and Loss Functions} \label{sec:three.four}

For pre-training on the source domain, we adopt the approach used in SHOT~\cite{liang2020really} to thoroughly capture the semantic information embedded in the source data. For fine-tuning in the target domain, as outlined in the above section, our objective is to ensure that the main target classifiers $C_{t_1}$ can correctly classify the target data. This objective is achieved through a two-step process.

\noindent\textbf{Step 1}: In the first step, we simultaneously train both the classifiers and the feature extractor. Since the support set in the few-shot setting is labeled, we can utilize the ground-truth labels directly, unlike traditional SFUDA methods that rely on pseudo-labels. To enable the feature extractor and classifiers to learn task-specific discriminative features, we minimize the label smoothing cross-entropy (LSCE) loss function during training.
	\begin{equation}
		\begin{aligned}
			\mathcal{L}_{lsce_1} = \mathcal{L}_{tar}^{l_t}(G_{t_1};\mathcal{X}_t,\mathcal{Y}_t) =& \\
			-\mathbb{E}_{(x_t,y_t)\in \mathcal{X}_t \times \mathcal{Y}_t}& \sum\nolimits_{k=1}^{K} q_k^{l_t} \log \delta_k(G_{t_1}(x_t)) \\
            \mathcal{L}_{lsce_2} = \mathcal{L}_{tar}^{l_t}(G_{t_2};\mathcal{X}_t,\mathcal{Y}_t) =& \\
			-\mathbb{E}_{(x_t,y_t)\in \mathcal{X}_t \times \mathcal{Y}_t}& \sum\nolimits_{k=1}^{K} q_k^{l_t} \log \delta_k(G_{t_2}(x_t))
		\end{aligned}
	\end{equation}
where $ \mathcal({X}_t,\mathcal{Y}_t) $ denotes support set and $(x_t,y_t)$ denotes labeled data in support set. 
$q^{l_t}_k=(1-\alpha)q_k + \alpha/K$ is the smoothed label and $\alpha$ is the smoothing parameter.
$ \delta_k(a)=\frac{\exp(a_k)}{\sum_i \exp(a_i)} $ denotes the $k$-th element in the soft-max output of a $K$-dimensional vector $a$.
$ G_{t_1} $, $ G_{t_2} $ denotes two branches of the target module of ACT, where $G_{t_i} = F_t + C_{t_i}, i= 1,2$. Optimizing the label smoothing cross-entropy loss function enhances the discriminability of the model and aids in aligning the target data more effectively by fully leveraging the semantic information present in the target domain.

Next, to enable the adapted model fit the target-specific structures well, we minimize the entropy of class-conditional distribution $r^t_j(\mathbf{x}^t_i) = p(y^t_i=j|\mathbf{x}^t_i;G_t)$ on target domain data $ \mathcal({X}_t,\mathcal{Y}_t) $ as
	\begin{equation}
		\begin{aligned}
			\mathcal{L}_{e_1} = \frac{1}{{{n_t}}}\sum\nolimits_{i = 1}^{{n_t}} {H( {G_{t_1}( {{\bf{x}}_i^t} )} )}\\
			\mathcal{L}_{e_2} = \frac{1}{{{n_t}}}\sum\nolimits_{i = 1}^{{n_t}} {H( {G_{t_2}( {{\bf{x}}_i^t} )} )}
		\end{aligned}
	\end{equation}
where $H(\cdot)$ denotes the entropy function of class-conditional distribution of two branches $G_{t_k}({{\bf{x}}_i^t}),~k=1,2$ defined as $ H({G_{t_k}({{\bf{x}}_i^t})}) = -\sum\nolimits_{j = 1}^c {{r^{t_k}_j}({{\bf{x}}_i^t})\log ({r^{t_k}_j}({{\bf{x}}_i^t})+\alpha)} $, $c$ is the number of classes, and ${r^{t_k}_j}\left( {{\mathbf{x}}_i^t} \right)$ is the probability of predicting point $\mathbf{x}_i^t$ to class $j$ for classifier $k$, $\alpha$ denotes a small parameter to avoid the situation that ${r^{t_k}_j} ({{\bf{x}}_i^t}) = 0$.

By applying entropy minimization, the two target classifiers, $c_{t_1}(\mathbf{x})$ and $c_{t_2}(\mathbf{x})$, are effectively aligned with the labeled support data and guided through the low-density regions of the target domain. This process helps the target model better capture and fit the underlying structures of the target data. 

Then, to minimize the discrepancy between source model output and adapted target model output, we train classifiers and feature extractor by minimizing the reverse cross-entropy loss function (RCE loss function):
    \begin{equation}
        \begin{aligned}
            \mathcal{L}_{rce_1} = -\sum\nolimits_{k=1}^{K} p_1(k|{\bf x}_i^t) \log (q_1(k|{\bf x}_i^t) + \alpha ) \\
            \mathcal{L}_{rce_2} = -\sum\nolimits_{k=1}^{K} p_2(k|{\bf x}_i^t) \log (q_2(k|{\bf x}_i^t) + \alpha )
        \end{aligned}     
    \end{equation}
where ${\bf x}_i^t$ denotes labeled support set data and $ q_1(k|{\bf x}_i^t) $, $ q_2(k|{\bf x}_i^t) $ denotes the outputs of the duplicate source classifiers $C_{s_1}$, $C_{s_2}$, respectively. 
$ p_1(k|{\bf x}_i^t) $, $ p_2(k|{\bf x}_i^t) $ denote the outputs of target classifiers $C_{t_1}$, $C_{t_2}$, respectively. 
$\alpha$ denotes a small parameter to avoid the situation that $ q(k|{\bf x}_i^t) = 0 $.

The RCE loss function is typically employed in scenarios with label noise, as it helps mitigate overfitting caused by noisy labels. In the few-shot setting, the support set contains only a small amount of ground-truth labeled data, which can lead to overfitting due to the limited labeled data availability. Furthermore, since we apply data augmentation to increase the data diversity, the augmented data may not perfectly align with the original ground-truth labels. By optimizing the RCE loss function, we not only prevent model collapse during training but also reduce overfitting in the support set by focusing on maintaining consistency between the data and labels. We combine the introduced objective functions to define the overall objective function of step A:
    \begin{equation}
        \begin{aligned}
               \mathop{\min}_{F_t,C_{t_1},C_{t_2}} \lambda_{lsce} (\mathcal{L}_{lsce_1} + \mathcal{L}_{lsce_2}) +\lambda_{e} (\mathcal{L}_{e_1} + \mathcal{L}_{e_2}) &\\
               + \lambda_{rce} (\mathcal{L}_{rce_1} + \mathcal{L}_{rce_2})
        \end{aligned}     
    \end{equation}
where $\lambda_{lsce}$, $\lambda_{e}$ and $\lambda_{rce}$ are hyperparameters that trader-off different objective functions.

\noindent\textbf{Step 2} In the second step, we train the classifiers $(C_{t_1},C_{t_2})$ while keeping the feature extractor $F_t$ fixed. The objective is to train the two distinct classifiers on the target domain to learn transferable features and establish a discriminative decision boundary specific to the target domain. To accomplish this, we employ the classifier determinacy disparity (CDD) distance, which measures the consistency and confidence of the outputs from both classifiers.

\begin{equation}
        \begin{aligned}
            \Gamma(\boldsymbol{p}_1,\boldsymbol{p}_2)= \sum_{m,n=1}^KA_{mn}-\sum_{m=1}^KA_{mm}=\sum_{m\neq n}^KA_{mn}
        \end{aligned}     
    \end{equation}
where $ p_1, p_2 \in \mathbb{R}^{K\times1}$ denotes the softmax probabilities of the outputs of the two classifiers and they need to satisfy:
    \begin{equation}
        \begin{aligned}
            \sum^K_{k=1}{p}_j^k = 1, ~\mathrm{s.t.} ~{p}_j^k \ge 0, ~\forall k = 1,\cdots, K, ~j = 1,2
        \end{aligned}     
    \end{equation}
$\mathbf{A}$ denotes a square matrix called Bi-classifier Prediction Relevance Matrix and $\mathbf{A} = \boldsymbol{p}_1\boldsymbol{p}^\top_2 \in \mathbb{R}^{K\times K}$.

The matrix $\mathbf{A}$ effectively assesses the predictive correlation between the two classifiers across different categories. To minimize the disparity in their prediction correlations, we aim to make the predictions from both classifiers consistent and well-aligned. This is achieved by maximizing the diagonal elements of $\mathbf{A}$, ensuring that the predicted class is assigned with high confidence. At the same time, the off-diagonal elements of $\mathbf{A}$ capture fine-grained confusion between the two classifiers. As a result, the CDD distance encapsulates all probabilities of inconsistency between the classifiers' predictions, serving as a comprehensive measure of the differences in their outputs.

We train the two classifiers on the target domain by maximizing CDD:
    \begin{equation}
        \begin{aligned}
            \mathcal{L}_{cdd} = \frac{1}{n_t}\sum^{n_t}_{i=1}\mathcal{L}_{cdd}({x}_i^t)
            =\frac{1}{n_t}\sum^{n_t}_{i=1}\Gamma({{p}_1}_i,{{p}_2}_i)\
        \end{aligned}     
    \end{equation}
where ${{p}_1}_i$, ${{p}_2}_i$ denotes the softmax outputs of $C_{t_1}$, $C_{t_2}$ for target sample ${x}_i^t$. 

We combine the introduced objective functions of the first step and introduced CDD distance in the second step to define the overall objective function of the second step:
    \begin{equation}
        \begin{aligned}
               \mathop{\min}_{C_{t_1},C_{t_2}} \lambda_{lsce} (\mathcal{L}_{lsce_1} + \mathcal{L}_{lsce_2}) +\lambda_{e} (\mathcal{L}_{e_1} + \mathcal{L}_{e_2}) &\\
               + \lambda_{rce} (\mathcal{L}_{rce_1} + \mathcal{L}_{rce_2}) - \lambda_{cdd} \mathcal{L}_{cdd}
        \end{aligned}     
    \end{equation}
where $\lambda_{lsce}$, $\lambda_{e}$ and $\lambda_{rce}$ are hyperparameters introduced in the first step and $\lambda_{cdd}$ is hyperparameter that trade-off CDD distance.
By optimizing this function, we can not only maintain the advantages of the first step, but also reduce the difference between the outputs of two classifiers. Both steps will be iteratively performed in our method.

\noindent\textbf{Optimizer} Traditional SFUDA methods typically rely on the SGD optimizer. In contrast, our approach employs a more advanced optimization technique known as Sharpness-Aware Minimization (SAM)~\cite{foret2020sharpness,kwon2021asam}. In essence, SAM performs a two-step optimization process, resetting the gradient to zero after each step. While this two-step approach slightly increases the training time, it significantly enhances the model's accuracy. In our experiments, we combined the SAM optimization method with the Adam optimizer, which yielded improved accuracy compared to the conventional SGD optimizer.

\section{Experiments}

In our experiments, we compare our ACT method with state-of-the-art SFUDA methods, as research on SFFSDA remains limited. Traditional SFUDA methods, such as~\cite{liang2020really,qiu2021source}, train the entire model on the labeled source domain, freezing certain components during adaptation to the target domain. Our approach utilizes a single base network and two classifiers. We believe the SFFSDA setting is more practical than the SFUDA setting as collecting limited amounts of labeled target data is more feasible than collecting large amounts of unlabeled target data. Given the few-shot setting, the support set often contains a limited amount of target data, resulting in insufficient semantic information. To address this limitation, we train the entire model on the labeled source domain and then fine-tune it on the labeled support set. Our proposed ACT method enables maximizing the utility of the semantic information contained in the support set.

\subsection{Setup}
\paragraph{Datasets.} We evaluate our method on four domain adaptation benchmarks to demonstrate its versatility.


\noindent \textbf{Office31}~\cite{saenko2010adapting} is a small benchmark with three domains: Amazon (A), DSLR (D), and Webcam (W), comprising 31 object categories in office environments.


\noindent \textbf{OfficeHome}~\cite{venkateswara2017deep} is a medium-sized benchmark with four domains: Art (Ar), Clipart (Cl), Product (Pr), and Real World (Rw), featuring 65 object categories.


\noindent \textbf{VisDA-C}~\cite{peng2017visda} is a large benchmark focused on synthesis-to-real object recognition, with 12 categories. The source domain includes 152,000 synthetic images, while the target domain contains 55,000 real-world images from the Microsoft COCO dataset.


\noindent \textbf{TerraIncognita (Terra)}\cite{beery2018recognition} contains wild animal images from four locations (L38, L43, L46, L100) across 10 categories. In the few-shot setting, cat and bird categories are often excluded due to limited samples.


We conduct single-source $ \rightarrow  $ single-target domain adaptation, averaging performance across all source $ \rightarrow  $ target pairs. To ensure reliability, we use three independently pre-trained source models (three different seeds) and a fixed seed for domain adaptation, enabling comparison with FT~\cite{lee2023fewshot}. Few-shot sample selection in the target domain is analyzed in the Appendix.

\paragraph{Baseline Method} We compare our ACT method with six SFUDA methods to demonstrate its effectiveness. The involved baselines are: SHOT~\cite{liang2020really}, SHOT++\cite{liang2021source}, AaD\cite{yang2022attracting}, G-SFDA~\cite{yang2021generalized}, NRC~\cite{yang2021exploiting}, and CoWA-JMDS~\cite{lee2022confidence}.  To ensure consistency, we take the convenience to reference results from ~\cite{lee2023fewshot}, which utilized a fixed random seed--a practice that mirrors our experimental setup--rather than citing outcomes from the original papers that were based on averages across three random runs.

\subsection{Implementation Details}
\paragraph{Network architecture.} We adopt the network designs proposed by~\cite{liang2020really}, a protocol widely embraced in SFUDA methods. Our architecture features a two branch design, wherein each branch integrates a backbone network along with two classifiers. In line with previous work~\cite{liang2020really,liang2021source,yang2021generalized,yang2021exploiting,lee2022confidence}, we employ ResNet-101 as the backbone for the VisDA-C dataset and opt for ResNet-50 across other benchmarks. Both backbone networks are pre-trained on ImageNet~\cite{deng2009imagenet}.

\paragraph{Network hyper-parameters.} For the source domain pre-training, the entire network is trained via back-propagation. For Office31 and OfficeHome, the learning rate is set to $ \eta_0 = 1\times e^{-3} $ for the backbone network and $ 10\times\eta_0 $ for the classifiers. For VisDA-C and Terra, $ \eta_0 = 1\times e^{-3} $ is applied to both. The pre-training process uses mini-batch SGD with a momentum of 0.9, weight decay of $5\times e^{-4}$, and batch sizes of 32 (Office31 and OfficeHome) and 128 (VisDA-C and Terra).

For the target domain fine-tuning, we employ the Adam optimizer with SAM. Batch sizes are set to 32 (Office31 and OfficeHome), 64 (VisDA-C), and 128 (Terra). The fine-tuning process uses 30,000 iterations, calculated as the product of the support set size and maximum epochs. For Office31, the learning rate is $ \eta_0 = 3\times e^{-5} $ for the base network and $ 10/3\times\eta_0 $ for the classifiers. Both learning rates follow the scheduler $\eta = \eta_0\times(1 + 10p)^{-0.75}$~\cite{ganin2015unsupervised,long2018conditional}, where $p$ represents the training progress $(0–1)$. For VisDA-C and Terra, we maintain the same learning rates as those utilized for Office31, with the learning rate scheduler being applied exclusively to the classifiers. For OfficeHome, a constant learning rate of $ \eta_0 = 1\times e^{-5} $ is applied on the base network and classifiers without a scheduler.



\subsection{Comparison with SFUDA Methods}\label{sec:four.three}

Firstly, we conduct a comparative analysis of our ACT method against several established SFUDA approaches. In our evaluation, we report the per-class average accuracy for the VisDA-C dataset and the overall average accuracy for other benchmarks. Within our SFFSDA setting, we employ 1, 3, 5, and 10 shots across all benchmarks with the exception of VisDA-C. Given its substantially larger scale, for VisDA-C, we increase the shot counts to 10, 20, 30, and 50 in our experiments.

As illustrated in Table~\ref{tab:one}, our method achieves superior performance relative to SFUDA approaches, even when utilizing only a limited amount of labeled data. Notably, for Terra, our method significantly outperforms all SFUDA methods across all few-shot settings. This exceptional performance is attributed to our ACT approach's effectiveness in mitigating the long-tail problem induced by uneven class distributions and ensuring balanced class representation through few-shot sampling. The label distribution for Terra is detailed in the Appendix. For VisDA-C, our method requires a 50-shot setting to match the competitive performance achieved by SFUDA baselines. In large-scale datasets such as VisDA-C, the number of labeled target instances used by our method is considerably smaller compared to the unlabeled data leveraged by SFUDA techniques. Specifically, 50 shots for 12 categories constitute a support set that amounts to approximately 1.08\% of the target domain data volume for VisDA-C. Although our ACT method does not surpass some SFUDA baselines on VisDA-C, it offers substantial resource savings during training by requiring only a limited amount of target data. Importantly, the 50-shot setting with our ACT method can yield competitive results with SHOT++, suggesting that a slight increase in data could lead to even better outcomes. Moreover, the performance gains observed in the 1-shot (or 10-shot for VisDA-C) settings markedly enhance the source pre-trained model's efficacy ("No adapt") across all datasets. These findings underscore that our ACT method not only addresses the long-tail distribution challenge commonly encountered in real-world applications but also delivers performance that rivals or exceeds that of SFUDA methods, while being more efficient in terms of data usage.

\begin{table}[t!]
    \begin{center}
    \resizebox{0.48\textwidth}{!}{%
    \begin{tabular}{lc|ccccc}
        \toprule
        \multicolumn{2}{l|}{\textbf{Method}} & \textbf{Office31} & \textbf{OfficeHome} & \textbf{VisDA-C}    & \textbf{Terra} \\
        \midrule
        \multicolumn{2}{l|}{No adapt}        & 77.63             & 59.26               & 47.78             & 29.18           \\
        \multicolumn{2}{l|}{SHOT}            & 87.79             & 69.90               & 81.77             & 25.72           \\
        \multicolumn{2}{l|}{SHOT$++$}        & 88.91             & 70.57               & \underline{86.84} & 27.07           \\
        \multicolumn{2}{l|}{AaD}             & 88.71             & 70.71               & 86.68             & 22.62            \\
        \multicolumn{2}{l|}{CoWA-JMDS}       & \underline{90.20} & \underline{72.07}   & 86.06             & 28.60            \\
        \multicolumn{2}{l|}{NRC}             & 88.59             & 70.25               & 85.31             & 26.35            \\
        \multicolumn{2}{l|}{G-SFDA}          & 86.99             & 69.22               & 84.72             & \underline{31.01}\\
        \midrule
        \multirow{4}{*}{ACT } 
        & $1$ ($10$)                         & 85.80             & 64.90               & 81.78             & \textbf{52.56}   \\ 
        & $3$ ($20$)                         & \textbf{91.27}    & \textbf{72.19}               & 84.18             & \textbf{55.75}   \\
        & $5$ ($30$)                         & \textbf{92.58}    & \textbf{75.75}      & 84.95             & \textbf{61.13}   \\
        & $10$ ($50$)                        & N/A               & \textbf{81.62}      & 86.44             & N/A              \\
        \bottomrule 
    \end{tabular}
    }
    \end{center}
    \caption{Classification accuracies (\%) of SFUDA and ACT methods. The $10$-shot setting is not considered in Office31 and Terra as some categories in these datasets have less than 10 images. \underline{Underline} indicates the best performance among SFUDA methods. \textbf{Bold} indicates the ACT's performance surpassing the SFUDA baselines.}
    \label{tab:one}
\end{table}

\subsection{Comparison with Transfer Learning Methods} 

\quad To substantiate the efficacy of transferring pre-trained source features to the target domain with minimal labeled data, we conduct a comparative analysis between our ACT method and several established transfer learning techniques in few-shot settings. A number of transfer learning methods have been proposed for few-shot scenarios, including L2-SP~\cite{xuhong2018explicit}, DELTA~\cite{li2019delta}, BSS~\cite{chen2019catastrophic}, and LCCS~\cite{zhang2022few}. To underscore the effectiveness of our ACT approach, we benchmark it against these methodologies on the Office31 and OfficeHome datasets, both under few-shot conditions. As summarized in Table~\ref{tab:two}, our method consistently outperforms the baseline transfer learning methods across all six evaluated settings. This superior performance not only highlights the advantage of our ACT method in few-shot scenarios but also underscores its capability to effectively leverage limited labeled data for successful domain adaptation. The results affirm that our ACT method represents a significant advancement over traditional transfer learning approaches in this context.

\quad 


\begin{table}[t!]
    \begin{center}
    \resizebox{0.48\textwidth}{!}{%
    \begin{tabular}{p{60pt}|ccc|ccc}
        \toprule
        & \multicolumn{3}{c|}{\textbf{Office31}} & \multicolumn{3}{c}{\textbf{OfficeHome}} \\
        & 1-shot & 3-shot & 5-shot & 1-shot & 3-shot & 5-shot \\
        \midrule
        L2-SP         & 82.99          & 87.43          & 90.20          & 64.00           & 70.13          & 73.34 \\
        DELTA         & 83.41          & 87.66          & 90.55          & 64.50           & 70.19          & 72.93 \\
        BSS           & 82.87          & 87.26          & 90.26          & 63.91           & 70.20          & 73.29 \\
        LCCS          & 79.16          & 85.75          & 89.17          & 60.53           & 65.54          & 68.18 \\
        \midrule
        ACT  & \textbf{85.80} & \textbf{91.27} & \textbf{92.58}          & \textbf{64.90}  & \textbf{72.19} & \textbf{75.75} \\
        \bottomrule
    \end{tabular}
    }
    \end{center}
    \caption{Classification accuracies (\%) for comparisons with transfer learning methods. \textbf{Bold} denotes the best performance for each few-shot setting.}
    \label{tab:two}
\end{table}

\subsection{Comparison with SFFSDA Method}

\quad Given that the SFFSDA setting is a nascent and emerging area, there has been limited exploration within this field. To date, only~\cite{lee2023fewshot} have conducted a preliminary investigation into SFFSDA, applying both fine-tuning (FT) and linear-probing fine-tuning (LP-FT) methods in this context. To provide deeper insights SFFSDA, we compare our ACT method against FT and LP-FT across all four benchmarks. As demonstrated in Table~\ref{tab:three}, our ACT method achieves superior performance across all few-shot settings, outperforming both FT and LP-FT approaches. Notably, on label-imbalanced datasets such as Office31 and Terra, our ACT method exhibits a more pronounced advantage over FT and LP-FT, highlighting its efficacy in addressing the long-tail distribution challenge. Furthermore, we have extended our experimental scope to two specialized DA scenarios: partial-set DA and open-set DA. The detailed findings from these extensive experiments are provided in the Appendix.


\begin{table}[t!]
    \begin{center}
    \resizebox{0.48\textwidth}{!}{%
    \begin{tabular}{lc|ccccc}
        \toprule
        \multicolumn{2}{l|}{\textbf{Method}} & \textbf{Office31} & \textbf{OfficeHome} & \textbf{VisDA-C}    & \textbf{Terra} \\
        \midrule
        \multirow{4}{*}{FT} 
        & $1$ ($10$)                         & 82.88             & 64.45               & 81.20             & 39.17            \\ 
        & $3$ ($20$)                         & 87.55             & 70.38               & 83.31             & 48.44            \\
        & $5$ ($30$)                         & 90.07             & 73.37               & 84.28             & 54.61            \\
        & $10$ ($50$)                        & N/A               & 76.99               & 85.93             & N/A              \\
        \midrule
        \multirow{4}{*}{LT-FT} 
        & $1$ ($10$)                         & 82.85             & 63.63               & 81.24             & 40.96            \\ 
        & $3$ ($20$)                         & 88.93             & 71.94               & 83.53             & 48.98            \\
        & $5$ ($30$)                         & 90.93             & 74.55               & 84.82             & 55.40            \\
        & $10$ ($50$)                        & N/A               & 78.58               & 86.17             & N/A              \\
        \midrule
        \multirow{4}{*}{ACT} 
        & $1$ ($10$)                         & \textbf{85.80}             & \textbf{64.90}               & \textbf{81.78}             & \textbf{52.56}            \\ 
        & $3$ ($20$)                         & \textbf{91.27}             & \textbf{72.19}               & \textbf{84.18}             & \textbf{55.75}            \\
        & $5$ ($30$)                         & \textbf{92.58}             & \textbf{75.75}               & \textbf{84.95}             & \textbf{61.13}            \\
        & $10$ ($50$)                        & N/A               & \textbf{81.62}               & \textbf{86.44}             & N/A              \\
        \bottomrule 
    \end{tabular}
    }
    \end{center}
    \caption{Classification accuracies (\%) of FT, LP-FT, and ACT methods on Office31, OfficeHome, VisDA-C, and Terra datasets. \textbf{Bold} denotes the best performance for each few-shot setting.}
    \label{tab:three}
\end{table}

\begin{table}[t!]
    \begin{center}
    \resizebox{0.48\textwidth}{!}{%
    \begin{tabular}{lcccccc}
        \toprule
         \begin{tabular}[c]{@{}c@{}}\textbf{cdd}\\ \textbf{distance}\end{tabular} & \begin{tabular}[c]{@{}c@{}c@{}}\textbf{source}\\ \textbf{model}\\ \textbf{matching}\end{tabular} & \begin{tabular}[c]{@{}c@{}c@{}}\textbf{strong}\\ \textbf{weak}\\ \textbf{augmentation}\end{tabular} & \begin{tabular}[c]{@{}c@{}}\textbf{SAM}\\ \textbf{optimization} \end{tabular} & \begin{tabular}[c]{@{}c@{}c@{}c@{}}\textbf{label}\\ \textbf{smooth}\\ \textbf{cross}\\ \textbf{entropy}\end{tabular} & \multicolumn{2}{|l}{\textbf{Acc (\%)}}  \\
        \midrule
        $\checkmark$          & $\checkmark$                  & $\checkmark$                        & $\checkmark$                  & $\checkmark$  & \multicolumn{2}{|l}{79.15}              \\
                               & $\checkmark$                  & $\checkmark$                        & $\checkmark$                  & $\checkmark$  & \multicolumn{2}{|l}{79.08}              \\
        $\checkmark$          &                               & $\checkmark$                        & $\checkmark$                  & $\checkmark$  & \multicolumn{2}{|l}{78.87}              \\
        $\checkmark$          & $\checkmark$                  &                                     & $\checkmark$                  & $\checkmark$  & \multicolumn{2}{|l}{78.48}              \\
        $\checkmark$          & $\checkmark$                  & $\checkmark$                        &                               & $\checkmark$  & \multicolumn{2}{|l}{78.23}              \\
        $\checkmark$          & $\checkmark$                  & $\checkmark$                        & $\checkmark$                  &               & \multicolumn{2}{|l}{76.70}              \\
        \bottomrule 
    \end{tabular}
    }
    \end{center}
    \caption{Ablation study results on Office31 dataset in 1-shot setting.}
    \label{tab:four}
\end{table}

\subsection{Ablation Study}

\quad We perform a 1-shot ablation study on the Office31 benchmark to isolate and evaluate the contributions of key components within our method:  CDD distance, source model matching, strong-weak augmentation, SAM optimization, and label-smooth cross-entropy. The findings are summarized in Table~\ref{tab:four}. Given the high degree of similarity between the Webcam (W) and DSLR (D) domains, we have omitted the results for D $\rightarrow$ W and W $\rightarrow$ D to focus more clearly on the impact of each individual component. As illustrated in Table~\ref{tab:four}, it is evident that each element plays a crucial role in enhancing model performance on the target domain.


\begin{itemize}  
    \item \textbf{CDD Distance:} Removing this component leads to a decrease in accuracy to 79.08\%, demonstrating the benefit of classifier matching in improving overall performance. 
    \item \textbf{Source Model Matching:} Omission of this component reduces accuracy to 78.87\%, highlighting the importance of retaining source model parameters in source-free settings. 
    \item \textbf{Strong-Weak Paradigm:} Eliminating this component lowers accuracy to 78.48\%, underscoring its effectiveness in improving data quality. 
    \item \textbf{SAM Optimization:} Excluding this component results in an accuracy drop to 78.23(\%), illustrating its role in enhancing the training process and optimizing over two stages. 
    \item \textbf{Label-Smooth Cross-Entropy:} Removing this component reduces performance to 76.70\%, indicating its contribution to enhancing the model's robustness when label information is available. 
\end{itemize}

\section{Conclusion}

In this work, we introduce the SFFSDA setting and propose the ACT method. To augment data diversity within the limited support set characteristic of few-shot learning, we have implemented a strong-weak augmentation paradigm, not only mitigating overfitting but also facilitating robust adaptation by using a minimal amount of labeled support data. Our ACT method is built on a two-step training process that integrates multiple loss functions, including label smoothing cross-entropy and classifier determinacy disparity. These losses enhance the discriminative power of the model while minimizing the discrepancy between the outputs of two classifiers. Furthermore, we employ SAM as our optimization strategy, which significantly contributes to the overall performance enhancement. The proposed framework enables effective adaptation of source models with as few as one sample per class, achieving results that are competitive with traditional SFUDA methods while overcoming their limitations. Our method demonstrates compelling performance across several DA benchmarks. Importantly, since acquiring a small amount of labeled target data is more practical than obtaining large volumes of unlabeled data in real-world applications, our ACT method presents a viable alternative to SFUDA approaches. Moreover, our study highlights the potential for further advancements in SFFSDA through the exploration of additional regularization techniques.


\bibliographystyle{named}
\bibliography{ijcai25}

\appendix

\section{Additional Results for Special Domain Adaptation Tasks}

Traditionally, domain adaptation (DA) research has extensively focused on closed-set scenarios~\cite{saenko2010adapting}, where the source and target domains share an identical label space. As DA research has progressed, two specialized settings have emerged to address more complex real-world challenges: open-set domain adaptation (ODA)~\cite{panareda2017open} and partial-set domain adaptation~\cite{cao2018partial}. In ODA, the target domain includes classes that are unseen during the training phase on the source domain, thereby requiring models to distinguish between known and unknown classes. Conversely, in PDA, the source domain's label space is broader than that of the target domain, necessitating methods that can effectively ignore extraneous classes present in the source but absent in the target domain. These advancements reflect a growing recognition of the need for DA techniques that can handle diverse and evolving label spaces, moving beyond the limitations of the traditional closed-set assumption.

For the PDA setting, we followed the protocol outlined in~\cite{liang2021source} using the OfficeHome dataset. Specifically, 25 classes (the first 25 alphabetically) out of 65 target domain classes were used. For the ODA setting, we adhered to the protocol in~\cite{liang2020really}, where the source domain contains 25 classes (the first 25 alphabetically), while the target domain includes all 65 classes, with some unknown samples. The empirical results of these experiments are presented in Table~\ref{tab:five}, where we compare our method against state-of-the-art techniques on the OfficeHome dataset under the 3-shot setting.

To understand why our method performs effectively in ODA and PDA scenarios, we begin by referencing SHOT~\cite{liang2021source}, the first SFUDA method. In PDA scenarios, SHOT utilizes self-supervised pseudo-labeling. Within this framework, although $K$ centroids are typically required, small or insignificant centroids should be treated as empty, akin to the approach used in k-means clustering. Specifically, SHOT discards centroids smaller than a predefined threshold $T_c$. In our FSSFDA setting, since we have access to the target labels, we can better adapt to the sparse target distribution, even though the target domain includes only a subset of the classes present in the source domain. So our method can work better than the traditional SFUDA method in the PDA scenario.

In the ODA setting, the target domain contains unknown classes that are absent in the source domain. To address this problem, in SHOT, the classifier layer remains unchanged, and a confidence thresholding strategy is employed to reject unknown samples during the learning process. Uncertainty is quantified using the entropy of the network output, normalized to the range [0,1], and is subsequently utilized for two-class k-means clustering in each epoch. Samples belonging to the cluster with higher uncertainty are treated as unknown and are excluded from target centroid updates and objective computations. In our ACT method, first, as we have access to target labels, we can ensure that the selected samples are sufficiently confident. Second, our approach computes the reverse cross-entropy loss between the source model's output and the adapted target model's output, thereby establishing a stronger relationship between the target and source domains. Consequently, our method outperforms traditional SFUDA methods in the ODA setting.

\subsection{Results of object recognition for PDA}


Compared to previous SFUDA methods, our approach achieves the highest average accuracy on the OfficeHome datasets. Specifically, our method surpasses the top-performing method, SHOT++, by 2.68\% in terms of average accuracy.

\subsection{Results of object recognition for ODA}

Compared with state-of-the-art SFUDA methods, our method obtains the best average accuracy on the OfficeHome datasets. Besides, our method outperforms the best method SHOT 1.51$\%$ in terms of average accuracy.


In summary, the results presented in Table ~\ref{tab:five} demonstrate that our method is both effective and competitive across these two challenging settings.

\begin{table*}[htbp]
    \begin{center}
	\resizebox{0.92\textwidth}{!}{$
	\begin{tabular}{lccccccccccccc}
		\toprule
		Partial-set DA (Source$\to$Target)  & Ar$\to$Cl & Ar$\to$Pr & Ar$\to$Re & Cl$\to$Ar & Cl$\to$Pr & Cl$\to$Re & Pr$\to$Ar & Pr$\to$Cl & Pr$\to$Re & Re$\to$Ar & Re$\to$Cl & Re$\to$Pr & Avg. \\
		\midrule
		ResNet-50 ~\cite{he2016deep} & 46.3 & 67.5 & 75.9 & 59.1 & 59.9 & 62.7 & 58.2 & 41.8 & 74.9 & 67.4 & 48.2 & 74.2 & 61.3 \\
		IWAN ~\cite{zhang2018importance}  & 53.9 & 54.5 & 78.1 & 61.3 & 48.0 & 63.3 & 54.2 & 52.0 & 81.3 & 76.5 & 56.8 & 82.9 & 63.6 \\
		SAN ~\cite{cao2018partial}  & 44.4 & 68.7 & 74.6 & 67.5 & 65.0 & 77.8 & 59.8 & 44.7 & 80.1 & 72.2 & 50.2 & 78.7 & 65.3 \\
		ETN ~\cite{cao2019learning}  & 59.2 & 77.0 & 79.5 & 62.9 & 65.7 & 75.0 & 68.3 & 55.4 & 84.4 & 75.7 & 57.7 & 84.5 & 70.5 \\
		SAFN ~\cite{xu2019larger}  & 58.9 & 76.3 & 81.4 & 70.4 & 73.0 & 77.8 & 72.4 & 55.3 & 80.4 & 75.8 & 60.4 & 79.9 & 71.8 \\
        DRCN ~\cite{li2020deep} & 54.0 & 76.4 & 83.0 & 62.1 & 64.5 & 71.0 & 70.8 & 49.8 & 80.5 & 77.5 & 59.1 & 79.9 & 69.0 \\
        RTNet$_{adv}$ ~\cite{chen2020selective} & 63.2 & 80.1 & 80.7 & 66.7 & 69.3 & 77.2 & 71.6 & 53.9 & 84.6 & 77.4 & 57.9 & 85.5 & 72.3 \\
        BA$^3$US ~\cite{liang2020balanced} & 60.6 & 83.2 & 88.4 & 71.8 & 72.8 & 83.4 & 75.5 & 61.6 & 86.5 & 79.3 & 62.8 & 86.1 & 76.0 \\
        TSCDA ~\cite{ren2020learning} & 63.6 & 82.5 & 89.6 & 73.7 & 73.9 & 81.4 & 75.4 & 61.6 & 87.9 & 83.6 & 67.2 & 88.8 & 77.4 \\
		\midrule
		SHOT-IM ~\cite{liang2020really} & 59.1 & 83.9 & 88.5 & 72.7 & 73.5 & 78.4 & 75.9 & 59.9 & 90.3 & 81.3 & 68.6 & 88.7 & 76.7 \\
		SHOT ~\cite{liang2020really} & 64.6 & 85.1 & 92.9 & 78.4 & 76.8 & 86.9 & 79.0 & 65.7 & 89.0 & 81.1 & 67.7 & 86.4 & 79.5 \\
        SHOT-IM++ ~\cite{liang2021source} & 59.6 & 84.5 & 89.0 & 73.7 & 74.2 & 79.3 & 77.0 & 60.7 & \textbf{91.0} & 81.8 & 69.4 & 89.3 & 77.5 \\
        SHOT++ ~\cite{liang2021source} & 65.0 & 85.8 & \textbf{93.4} & 78.8 & 77.4 & \textbf{87.3} & 79.3 & 66.0 & 89.6 & 81.3 & 68.1 & 86.8 & 79.9 \\
        \midrule
        ACT & \textbf{70.73} & \textbf{89.48} & 90.65 & \textbf{79.65} & \textbf{86.22} & 86.44 & \textbf{80.80} & \textbf{71.08} & 89.15 & \textbf{83.96} & \textbf{72.50} & \textbf{90.29} & \textbf{82.58} \\
		\midrule\midrule
		Open-set DA (Source$\to$Target)  & Ar$\to$Cl & Ar$\to$Pr & Ar$\to$Re & Cl$\to$Ar & Cl$\to$Pr & Cl$\to$Re & Pr$\to$Ar & Pr$\to$Cl & Pr$\to$Re & Re$\to$Ar & Re$\to$Cl & Re$\to$Pr & Avg. \\
		\midrule
		ResNet-50 ~\cite{he2016deep}   & 53.4 & 52.7 & 51.9 & 69.3 & 61.8 & 74.1 & 61.4 & 64.0 & 70.0 & 78.7 & 71.0 & 74.9 & 65.3 \\
		ATI-$\lambda$ ~\cite{panareda2017open}  & 55.2 & 52.6 & 53.5 & 69.1 & 63.5 & 74.1 & 61.7 & 64.5 & 70.7 & 79.2 & 72.9 & 75.8 & 66.1 \\
		OSBP ~\cite{saito2018open}    & 56.7 & 51.5 & 49.2 & 67.5 & 65.5 & 74.0 & 62.5 & 64.8 & 69.3 & 80.6 & 74.7 & 71.5 & 65.7 \\
		OpenMax ~\cite{bendale2016towards} & 56.5 & 52.9 & 53.7 & 69.1 & 64.8 & 74.5 & 64.1 & 64.0 & 71.2 & 80.3 & 73.0 & 76.9 & 66.7 \\
		STA ~\cite{liu2019separate}    & 58.1 & 53.1 & 54.4 & 71.6 & 69.3 & 81.9 & 63.4 &65.2 & 74.9 & 85.0 & 75.8 & 80.8 & 69.5 \\
		\midrule
		SHOT-IM ~\cite{liang2020really} & 62.5 & 77.8 & 83.9 & 60.9 & 73.4 & 79.4 & 64.7 & 58.7 & 83.1 & 69.1 & 62.0 & 82.1 & 71.5 \\
		SHOT ~\cite{liang2020really} & \textbf{64.5} & 80.4 & \textbf{84.7} & 63.1 & 75.4 & \textbf{81.2} & 65.3 & 59.3 & \textbf{83.3} & 69.6 & \textbf{64.6} & 82.3 & 72.8 \\
		\midrule
        ACT & 63.62 & \textbf{82.98} & 83.80 & \textbf{67.85} & \textbf{77.78} & 79.79 & \textbf{68.48} & \textbf{61.40} & 81.71 & \textbf{75.02} & 64.25 & \textbf{85.09} & \textbf{74.31} \\
        \midrule
		\end{tabular}
		$}
        \end{center}
        \caption{Classification accuracies (\%) on OfficeHome dataset for partial-set and open-set DA (3 shot). \textbf{Bold} denotes the best performance.}
        \label{tab:five}
\end{table*} 

\begin{table*}[htbp]
    \begin{center}
	\resizebox{0.92\textwidth}{!}{$
	\begin{tabular}{lc|cccccccc}
		\toprule
		Method & & \multicolumn{1}{c}{A→D} & \multicolumn{1}{c}{A→W} & \multicolumn{1}{c}{D→A} & \multicolumn{1}{c}{D→W} & \multicolumn{1}{c}{W→A} & \multicolumn{1}{c}{W→D} & \multicolumn{1}{c}{Avg}\\
		\midrule
		\multicolumn{2}{l|}{No adapt} & $77.33$ & $76.52$ & $57.33$ & $93.92$ & $62.00$ & $98.67$ & $77.63$ \\
        \multicolumn{2}{l|}{SHOT~\cite{liang2020really}} & $89.33$ & $92.66$ & $73.82$ & $98.32$ & $72.64$ & $100.00$ & $87.79$ \\
        \multicolumn{2}{l|}{SHOT$++$~\cite{liang2021source}} & $93.67$ & $92.45$ & $74.76$ & $98.11$ & $75.47$ & $99.00$ & $88.91$ \\
        \multicolumn{2}{l|}{AaD~\cite{yang2022attracting}} & $93.33$ & $93.08$ & $73.94$ & $97.90$ & $74.35$ & $99.67$ & $88.71$ \\
        \multicolumn{2}{l|}{CoWA-JMDS~\cite{lee2022confidence}} & $94.67$ & $94.97$ & $76.00$ & $98.53$ & $77.37$ & $99.67$ & $\underline{90.20}$ \\
        \multicolumn{2}{l|}{NRC~\cite{yang2021exploiting}} & $91.67$ & $91.82$ & $74.94$ & $98.11$ & $75.65$ & $99.33$ & $88.59$ \\
        \multicolumn{2}{l|}{G-SFDA~\cite{yang2021generalized}} & $88.67$ & $89.52$ & $73.47$ & $97.27$ & $72.99$ & $100.00$ & $86.99$ \\
        \midrule
        \multirow{3}{*}{ACT} & $1$ & $93.22$ & $85.64$ & $68.09$ & $98.25$ & $69.65$ & $99.93$ & $85.80$ \\ %
        & $3$ & $97.70$ & $95.73$ & $77.08$ & $99.67$ & $77.46$ & $100.00$ & $\bold{91.27}$ \\
        & $5$ & $99.13$ & $97.24$ & $79.81$ & $99.32$ & $79.95$ & $100.00$ & $\bold{92.58}$ \\
        \midrule
		\end{tabular}
		$}
        \end{center}
        \caption{Accuracies of SFUDA and ACT method on the Office31 dataset. A, D, and W denote Amazon, DSLR, and Webcam, respectively. The numbers next to ACT represent the number of labeled images per class used for training. \underline{Underline} indicates the best performance among SFUDA methods. \textbf{Bold} indicate fine-tuning performance surpassing the best performance of SFUDA methods.}
        \label{tab:six}
\end{table*} 

\begin{table*}[htbp]
    \begin{center}
	\resizebox{0.92\textwidth}{!}{$
	\begin{tabular}{lc|cccccccccccccc}
		\toprule
        Method & & \multicolumn{1}{c}{Ar→Cl} & \multicolumn{1}{c}{Ar→Pr} & \multicolumn{1}{c}{Ar→Rw} & \multicolumn{1}{c}{Cl→Ar} & \multicolumn{1}{c}{Cl→Pr} & \multicolumn{1}{c}{Cl→Rw} & \multicolumn{1}{c}{Pr→Ar} & \multicolumn{1}{c}{Pr→Cl} & \multicolumn{1}{c}{Pr→Rw} & \multicolumn{1}{c}{Rw→Ar} & \multicolumn{1}{c}{Rw→Cl} & \multicolumn{1}{c}{Rw→Pr} & \multicolumn{1}{c}{Avg} \\
		\midrule
		\multicolumn{2}{l|}{No adapt} & $48.19$ & $64.56$ & $73.20$ & $49.93$ & $61.64$ & $62.16$ & $49.79$ & $42.31$ & $72.21$ & $61.59$ & $48.30$ & $77.25$ & $59.26$ \\
        \multicolumn{2}{l|}{SHOT~\cite{liang2020really}} & $54.91$ & $75.45$ & $79.97$ & $64.68$ & $75.60$ & $77.14$ & $64.61$ & $53.68$ & $81.38$ & $71.74$ & $57.39$ & $82.28$ & $69.90$ \\
        \multicolumn{2}{l|}{SHOT$++$~\cite{liang2021source}} & $56.05$ & $77.78$ & $80.12$ & $65.57$ & $78.90$ & $77.37$ & $65.30$ & $53.61$ & $80.70$ & $71.33$ & $56.51$ & $83.56$ & $70.57$ \\
        \multicolumn{2}{l|}{AaD~\cite{yang2022attracting}} & $56.16$ & $77.89$ & $79.74$ & $63.72$ & $79.99$ & $78.52$ & $63.10$ & $56.13$ & $80.81$ & $69.75$ & $57.85$ & $84.83$ & $70.71$ \\
        \multicolumn{2}{l|}{CoWA-JMDS~\cite{lee2022confidence}} & $57.47$ & $78.60$ & $81.50$ & $67.76$ & $79.43$ & $79.89$ & $65.77$ & $56.74$ & $82.27$ & $71.40$ & $59.91$ & $84.08$ & $\underline{72.07}$ \\
        \multicolumn{2}{l|}{NRC~\cite{yang2021exploiting}} & $55.59$ & $76.76$ & $79.74$ & $65.50$ & $76.80$ & $77.41$ & $65.09$ & $54.94$ & $80.51$ & $71.12$ & $56.97$ & $82.51$ & $70.25$ \\
        \multicolumn{2}{l|}{G-SFDA~\cite{yang2021generalized}} & $54.98$ & $75.19$ & $79.82$ & $64.27$ & $73.65$ & $75.88$ & $63.44$ & $52.58$ & $80.58$ & $70.30$ & $57.58$ & $82.36$ & $69.22$ \\
        \midrule
        \multirow{4}{*}{ACT} & $1$ & $51.86$ & $69.78$ & $73.39$ & $56.86$ & $68.28$ & $67.91$ & $58.82$ & $51.09$ & $75.07$ & $69.11$ & $56.77$ & $79.83$ & $64.90$ \\ %
        & $3$ & $61.47$ & $77.89$ & $77.18$ & $66.66$ & $78.27$ & $74.48$ & $67.74$ & $61.75$ & $77.74$ & $74.22$ & $64.49$ & $84.35$ & $\bold{72.19}$ \\
        & $5$ & $66.89$ & $81.77$ & $79.37$ & $71.42$ & $81.45$ & $74.95$ & $72.15$ & $67.21$ & $81.42$ & $77.02$ & $68.99$ & $86.32$ & $\bold{75.75}$\\
        & $10$ & $74.37$ & $87.34$ & $84.44$ & $78.93$ & $86.44$ & $81.22$ & $80.04$ & $74.31$ & $84.24$ & $82.61$ & $75.76$ & $89.74$ & $\bold{81.62}$ \\
        \midrule
		\end{tabular}
		$}
        \end{center}
        \caption{Accuracies of SFUDA and ACT method on the OfficeHome dataset. Ar, Cl, Pr, and Rw denote Art, Clipart, Product, and Real World, respectively. Other notations are the same as in Table~\ref{tab:six}.}
        \label{tab:seven}
\end{table*} 

\begin{table*}[htbp]
    \begin{center}
	\resizebox{0.92\textwidth}{!}{$
	\begin{tabular}{lc|cccccccccccccc}
		\toprule
        Method & & \multicolumn{1}{c}{aeroplane} & \multicolumn{1}{c}{bicycle} & \multicolumn{1}{c}{bus} & \multicolumn{1}{c}{car} & \multicolumn{1}{c}{horse} & \multicolumn{1}{c}{knife} & \multicolumn{1}{c}{motorcycle} & \multicolumn{1}{c}{person} & \multicolumn{1}{c}{plant} & \multicolumn{1}{c}{skateboard} & \multicolumn{1}{c}{train} & \multicolumn{1}{c}{truck} & \multicolumn{1}{c}{Avg}\\
		\midrule
		\multicolumn{2}{l|}{No adapt} & $60.62$ & $19.60$ & $52.24$ & $73.26$ & $64.99$ & $4.45$ & $82.75$ & $16.63$ & $73.15$ & $35.66$ & $82.46$ & $7.53$ & $47.78$ \\
        \multicolumn{2}{l|}{SHOT~\cite{liang2020really}} & $95.16$ & $89.06$ & $79.39$ & $53.66$ & $92.79$ & $94.38$ & $79.38$ & $80.83$ & $89.49$ & $86.62$ & $86.74$ & $53.72$ & $81.77$ \\
        \multicolumn{2}{l|}{SHOT$++$~\cite{liang2021source}} & $97.21$ & $86.42$ & $89.48$ & $84.93$ & $97.86$ & $97.51$ & $92.66$ & $84.00$ & $96.96$ & $92.47$ & $93.55$ & $29.04$ & $\underline{86.84}$ \\
        \multicolumn{2}{l|}{AaD~\cite{yang2022attracting}} & $96.43$ & $91.02$ & $88.95$ & $80.40$ & $96.56$ & $96.64$ & $89.71$ & $79.83$ & $94.39$ & $90.87$ & $91.01$ & $47.71$ & $86.68$ \\
        \multicolumn{2}{l|}{CoWA-JMDS~\cite{lee2022confidence}} & $95.33$ & $88.49$ & $82.12$ & $71.21$ & $95.70$ & $97.99$ & $89.79$ & $85.54$ & $92.82$ & $90.35$ & $87.29$ & $56.07$ & $86.06$ \\
        \multicolumn{2}{l|}{NRC~\cite{yang2021exploiting}} & $96.98$ & $93.19$ & $83.97$ & $60.60$ & $96.23$ & $95.50$ & $81.48$ & $79.75$ & $93.56$ & $91.96$ & $90.67$ & $59.79$ & $85.31$ \\
        \multicolumn{2}{l|}{G-SFDA~\cite{yang2021generalized}} & $96.48$ & $88.20$ & $84.40$ & $70.64$ & $95.92$ & $96.38$ & $89.04$ & $79.46$ & $93.78$ & $91.01$ & $89.22$ & $42.07$ & $84.72$ \\
        \midrule
        \multirow{4}{*}{ACT} & $10$ & $91.59$ & $81.89$ & $79.40$ & $75.07$ & $94.33$ & $90.61$ & $81.92$ & $84.74$ & $91.10$ & $79.86$ & $83.68$ & $47.20$ & $81.78$ \\ 
        & $20$ & $93.79$ & $85.53$ & $82.24$ & $74.00$ & $94.18$ & $89.37$ & $84.46$ & $87.22$ & $90.79$ & $89.12$ & $83.68$ & $55.73$ & $84.18$ \\
        & $30$ & $94.69$ & $88.15$ & $77.15$ & $71.25$ & $94.31$ & $91.94$ & $85.38$ & $89.00$ & $92.30$ & $89.85$ & $86.62$ & $58.69$ & $84.95$ \\
        & $50$ & $95.89$ & $86.77$ & $82.36$ & $75.10$ & $94.69$ & $93.64$ & $88.26$ & $86.60$ & $92.53$ & $92.94$ & $86.83$ & $61.60$ & $86.44$ \\
        \midrule
		\end{tabular}
		$}
        \end{center}
        \caption{Accuracies of SFUDA and ACT method on the VisDA-C dataset (Synthesis $\rightarrow$ Real). Other notations are the same as in Table~\ref{tab:six}.}
        \label{tab:eight}
\end{table*} 

\begin{table*}[htbp]
    \begin{center}
	\resizebox{0.92\textwidth}{!}{$
	\begin{tabular}{lc|cccccccccccccc}
		\toprule
        Method & & \multicolumn{1}{c}{L100$\rightarrow$L38} & \multicolumn{1}{c}{L100$\rightarrow$L43} & \multicolumn{1}{c}{L100$\rightarrow$L46} & \multicolumn{1}{c}{L38$\rightarrow$L100} & \multicolumn{1}{c}{L38$\rightarrow$L43} & \multicolumn{1}{c}{L38$\rightarrow$L46} & \multicolumn{1}{c}{L43$\rightarrow$L100} & \multicolumn{1}{c}{L43$\rightarrow$L38} & \multicolumn{1}{c}{L43$\rightarrow$L46} & \multicolumn{1}{c}{L46$\rightarrow$L100} & \multicolumn{1}{c}{L46$\rightarrow$L38} & \multicolumn{1}{c}{L46$\rightarrow$L43} & \multicolumn{1}{c}{Avg}\\
		\midrule
		\multicolumn{2}{l|}{No adapt} & $26.24$ & $20.34$ & $27.08$ & $29.31$ & $31.35$ & $31.59$ & $24.11$ & $44.09$ & $38.66$ & $33.60$ & $21.57$ & $22.18$ & $29.18$ \\
        \multicolumn{2}{l|}{SHOT~\cite{liang2020really}} & $20.07$ & $23.82$ & $28.48$ & $35.95$ & $28.98$ & $13.57$ & $26.20$ & $14.52$ & $32.72$ & $34.32$ & $12.63$ & $37.41$ & $25.72$ \\
        \multicolumn{2}{l|}{SHOT$++$~\cite{liang2021source}} & $29.32$ & $22.08$ & $25.47$ & $22.76$ & $31.79$ & $18.44$ & $33.31$ & $22.61$ & $25.60$ & $35.82$ & $13.00$ & $44.59$ & $27.07$ \\
        \multicolumn{2}{l|}{AaD~\cite{yang2022attracting}} & $17.15$ & $17.35$ & $22.14$ & $24.56$ & $28.10$ & $13.30$ & $28.92$ & $23.34$ & $23.06$ & $31.56$ & $7.35$ & $34.59$ & $22.62$ \\
        \multicolumn{2}{l|}{CoWA-JMDS~\cite{lee2022confidence}} & $33.08$ & $31.42$ & $26.35$ & $36.26$ & $38.33$ & $19.25$ & $28.23$ & $13.57$ & $26.62$ & $32.52$ & $9.96$ & $47.56$ & $28.60$ \\
        \multicolumn{2}{l|}{NRC~\cite{yang2021exploiting}} & $19.28$ & $22.70$ & $29.47$ & $38.46$ & $26.88$ & $14.94$ & $30.78$ & $22.64$ & $32.16$ & $28.86$ & $11.04$ & $39.04$ & $26.35$ \\
        \multicolumn{2}{l|}{G-SFDA~\cite{yang2021generalized}} & $21.59$ & $29.14$ & $38.17$ & $38.38$ & $27.03$ & $22.40$ & $40.86$ & $17.42$ & $33.29$ & $35.01$ & $16.25$ & $52.55$ & $\underline{31.01}$ \\
        \midrule
        \multirow{3}{*}{ACT} & $1$ & $63.73$ & $36.38$ & $40.20$ & $55.79$ & $40.65$ & $32.69$ & $60.57$ & $62.41$ & $46.00$ & $68.70$ & $64.74$ & $58.81$ & $\bold{52.56}$ \\ %
        & $3$ & $65.14$ & $38.92$ & $44.04$ & $62.46$ & $42.76$ & $38.79$ & $69.71$ & $66.92$ & $45.03$ & $68.22$ & $67.44$ & $59.54$ & $\bold{55.75}$ \\
        & $5$ & $66.79$ & $47.16$ & $51.05$ & $68.57$ & $48.94$ & $50.93$ & $76.00$ & $68.09$ & $58.38$ & $71.15$ & $66.42$ & $60.09$ & $\bold{61.13}$ \\
        \midrule
		\end{tabular}
		$}
        \end{center}
        \caption{Accuracies of SFUDA and ACT method on the Terra dataset. L100, L38, L43, and L46 denote different locations (i.e., domains). Other notations are the same as in Table~\ref{tab:six}.}
        \label{tab:nine}
\end{table*} 

\begin{table*}[htbp]
    \begin{center}
	\resizebox{0.92\textwidth}{!}{$
	\begin{tabular}{lc|cccccccccccccc}
		\toprule
        Method & & \multicolumn{1}{c}{Ar→Cl} & \multicolumn{1}{c}{Ar→Pr} & \multicolumn{1}{c}{Ar→Rw} & \multicolumn{1}{c}{Cl→Ar} & \multicolumn{1}{c}{Cl→Pr} & \multicolumn{1}{c}{Cl→Rw} & \multicolumn{1}{c}{Pr→Ar} & \multicolumn{1}{c}{Pr→Cl} & \multicolumn{1}{c}{Pr→Rw} & \multicolumn{1}{c}{Rw→Ar} & \multicolumn{1}{c}{Rw→Cl} & \multicolumn{1}{c}{Rw→Pr} & \multicolumn{1}{c}{Avg} \\
		\midrule
        \multirow{4}{*}{ACT} & $2019$ & $66.38$ & $81.11$ & $80.34$ & $68.90$ & $80.13$ & $75.82$ & $71.64$ & $66.65$ & $80.86$ & $76.83$ & $69.13$ & $86.16$ & $75.33$ \\ %
        & $2020$ & $65.41$ & $82.17$ & $80.58$ & $71.63$ & $80.34$ & $76.76$ & $72.69$ & $66.61$ & $81.37$ & $77.99$ & $68.86$ & $86.70$ & $75.93$ \\
        & $2021$ & $65.18$ & $82.01$ & $80.79$ & $70.60$ & $80.53$ & $76.41$ & $71.74$ & $67.59$ & $80.82$ & $76.97$ & $69.94$ & $86.14$ & $75.73$\\
        \midrule
		\end{tabular}
		$}
        \end{center}
        \caption{Accuracies of ACT method on the OfficeHome dataset using different target random seed. Ar, Cl, Pr, and Rw denote Art, Clipart, Product, and Real World, respectively. The numbers next to ACT represent the number of random seeds.}
        \label{tab:ten}
\end{table*} 

\section{Label Distributions of Datasets}

\begin{figure}[!h]
    \centering
    \includegraphics[width=0.9\columnwidth]{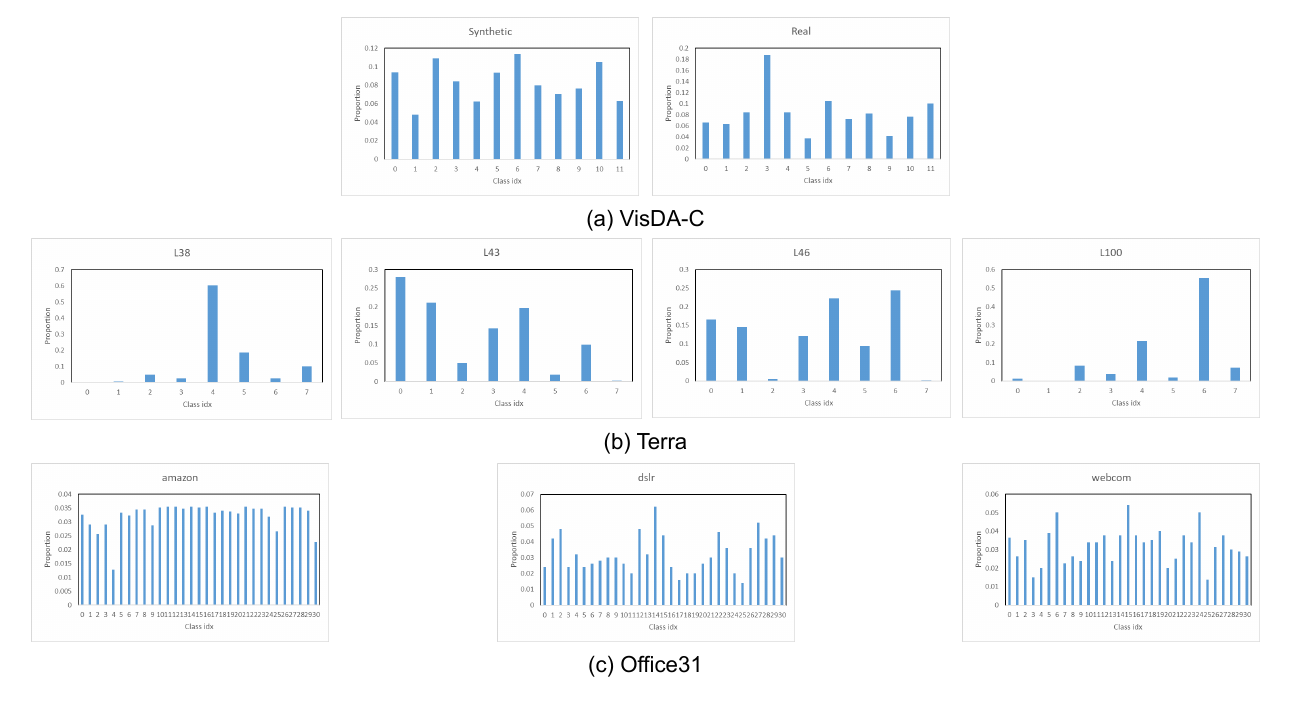}
    \caption{The label distributions of the three benchmarks: VisDA-C, Terra, and Office31. In each histogram, the x-axes and y-axes represent the class index and the proportion of this class in the domain, respectively}
    \label{fig3}
\end{figure}

Figure~\ref{fig3} illustrates the label distribution for each domain across the three benchmarks used in this study. Office31, one of the most widely used benchmarks in the SFUDA literature, exhibits relatively balanced label distributions across all domains. In contrast, the Terra datasets show significant shifts in label distributions.

\section{Qualitative Results of Experiments}

\begin{figure}[H]
    \centering
    \includegraphics[width=0.9\columnwidth]{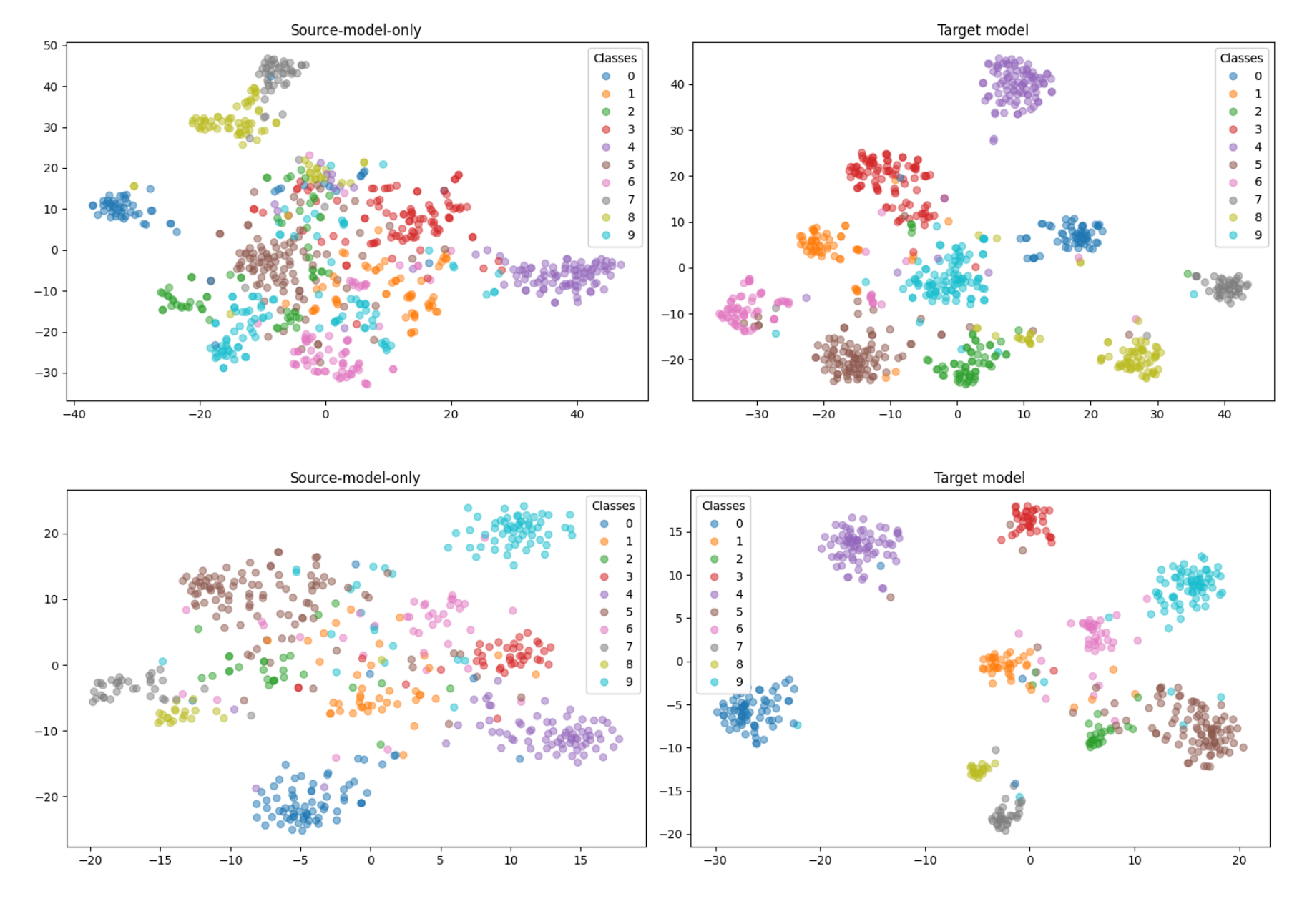}
    \caption{$t$-SNE visualization results on the Office-Home dataset (Ar $\rightarrow$ Cl(top) and Cl $\rightarrow$ Ar(bottom)). The left picture indicates the result from the source pre-trained model. The right picture indicates the result from the adapted model. Different colors represent different classes. The first $10$ classes are sampled for clear visualization.}
    \label{fig:tsne1}
\end{figure}
\begin{figure}[H]
    \centering
    \includegraphics[width=0.9\columnwidth]{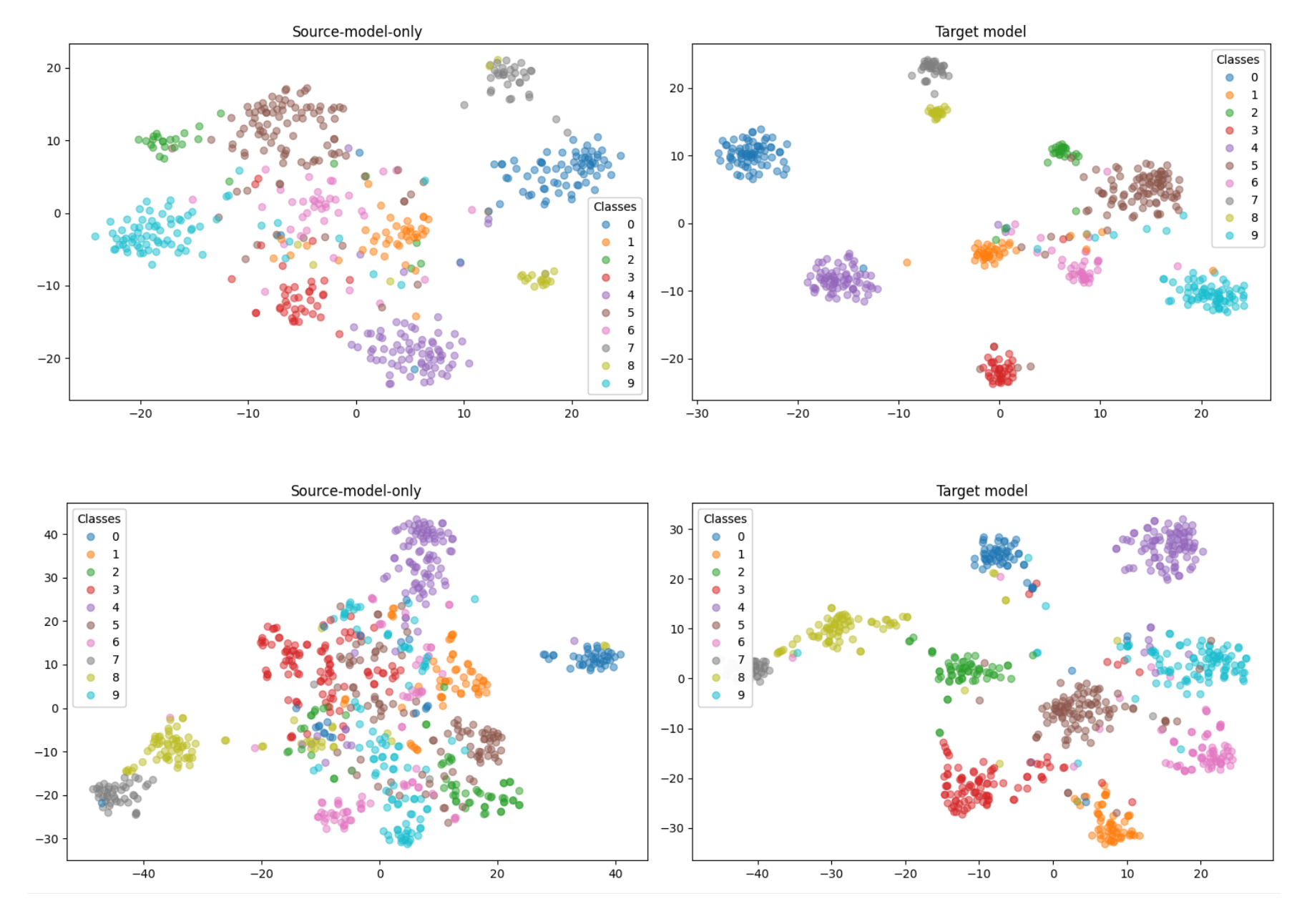}
    \caption{$t$-SNE visualization results on the Office-Home dataset (Ar $\rightarrow$ Cl(top) and Cl $\rightarrow$ Ar(bottom)). The left picture indicates the result from the source pre-trained model. The right picture indicates the result from the adapted model. Different colors represent different classes. The first $10$ classes are sampled for clear visualization.}
    \label{fig:tsne2}
\end{figure}

We present several qualitative results using $t$-SNE~\cite{van2008visualizing} visualizations on the OfficeHome dataset in four source $\rightarrow$ target tasks: Ar $\rightarrow$ Cl, Cl $\rightarrow$ Ar, Pr $\rightarrow$ Ar and Rw $\rightarrow$ Cl. The analysis includes the source pre-trained model both before and after adaptation to the target domain with 10-shot samples. As shown in Figure~\ref{fig:tsne1} and Figure~\ref{fig:tsne2}, the source pre-trained model, without target data adaptation, struggles to effectively distinguish between target data, indicating poor classification performance on the target domain. However, after adaptation, the target model significantly improves in its ability to discriminate between target data, demonstrating that our method enhances the discriminative power of the adapted model.

\section{Discussion of Training Process and Time Consumption}
Our method involves a complex training process with two-step optimization and multiple loss functions, which require slightly more time to execute. However, in a few-shot setting, the primary goal is to minimize the target dataset size, making training time less critical. Additionally, training time and model performance can be balanced by reducing the number of iterations, implementing early stopping when performance plateaus, and similar strategies.

\section{Analysis of the Influence of the Quality of the Chosen Few-shot Samples}
Since our method utilizes data from the target domain through random few-shot sampling, the quality of the selected few-shot samples plays a crucial role in the few-shot setting. To evaluate this factor, we conducted experiments using three random seeds for data sampling in the target domain. Additionally, we employed three source pre-trained models, consistent with those used in other experiments. The results, presented in ~\ref{tab:ten}, reveal that the difference between the highest and lowest performance is 0.6\%, which is less than 1\%. Furthermore, the variance among the three results is less than 0.5\%. These findings indicate that our method effectively mitigates the impact of the quality of selected few-shot samples in the target domain, demonstrating its potential for widespread application in various real-world scenarios.

\end{document}